# Hilbert Space Embeddings of POMDPs

**Yu Nishiyama**[1] **Abdeslam Boularias**[2] **Arthur Gretton**[2,3] **Kenji Fukumizu**[1]
[1]The Institute of Statistical Mathematics, {nishiyam,fukumizu}@ism.ac.jp
[2]Max Planck Institute for Intelligent Systems, boularias@tuebingen.mpg.de
[3]Gatsby Computational Neuroscience Unit, CSML, UCL, arthur.gretton@gmail.com

## Abstract

A nonparametric approach for policy learning for POMDPs is proposed. The approach represents distributions over the states, observations, and actions as embeddings in feature spaces, which are reproducing kernel Hilbert spaces. Distributions over states given the observations are obtained by applying the kernel Bayes' rule to these distribution embeddings. Policies and value functions are defined on the feature space over states, which leads to a feature space expression for the Bellman equation. Value iteration may then be used to estimate the optimal value function and associated policy. Experimental results confirm that the correct policy is learned using the feature space representation.

## 1 Introduction

Partially Observable Markov Decision Processes (POMDPs) are general models for sequential control problems in partially observable environments, in which an agent executes an action under uncertainty while reward delivery and state transitions vary depending on the state and action. The objective is to find an optimal policy that maximizes a value function defined on beliefs (distributions over states), and determined by the reward function. When the value is the expected sum of discounted rewards, the optimal policy and the optimal value function are computed by solving a Bellman equation. Solutions to the Bellman equation are generally very difficult to obtain, with a number of approaches being employed. These include tractable parametric models of the system (parametric POMDPs [Poupart et al., 2006, Even-dar, 2005]); Monte Carlo methods for more complex models (Monte Carlo POMDPs [Silver and Veness, 2010, Thrun, 2000]); and methods that take advantage of the piece-wise linear and convex (PWLC) property (PBVI [Pineau et al., 2003], Perseus [Spaan and Vlassis, 2005], HSVI [Smith and Simmons, 2004]), where these last approaches use the fact that value functions computed by finite-step value iteration algorithms are PWLC in the beliefs [Sondik, 1971, Porta. et al., 2006]. All these methods have drawbacks, however: parametric models can cause bias errors if they oversimplify, Monte Carlo sampling methods can be computationally costly, and PWLC-based methods require the observations and actions to be discrete.

Our nonparametric approach is based on a series of recent works in which probability distributions are represented as embeddings in reproducing kernel Hilbert spaces (RKHSs) [Fukumizu et al., 2008, Gretton et al., 2007, Gretton et al., 2012]. Probability distributions can be embedded as points in RKHSs, and expectations of RKHS functions may be obtained as inner products with these embeddings, using the kernel trick. For a sufficiently rich RKHS (i.e., a characteristic RKHS), every probability distribution has a unique embedding, and the distance between embeddings is a metric on distributions [Fukumizu et al., 2008, Sriperumbudur et al., 2010]. Recently, inference methods which make use of the embeddings have been proposed, including for Hidden Markov Models [Song et al., 2009, Song et al., 2010], and belief propagation [Song et al., 2010, Song et al., 2011]. More recently, the embeddings have been used in performing value iteration and optimal policy learning in Markov decision processes (MDPs) [Grunewalder et al., 2012]; and the Kernel Bayes' Rule (KBR) [Fukumizu et al., 2011] was proposed, which updates a prior distribution embedding via feature space operations to obtain a posterior embedding.

In this paper, we propose kernel POMDPs (kPOMDPs), a new approach to solving the Bellman equations and determining policy, based on a non-parametric model defined in appropriate RKHSs. All probability distributions required for the algorithm are represented as embeddings in RKHSs, including the beliefs over the states, the transition models, the observation models, and the predictive distributions over subsequent states given actions, observations, and current beliefs. These embeddings are updated sequentially based on actions and observations. Likewise, value functions are defined over feature representations of states, and policies map RKHS representations of states to actions, which leads to an expression for the Bellman equation in feature space.

We use the feature representation of the Bellman equations to define a value iteration algorithm for POMDPs, which directly estimates both the expected immediate rewards and expected values of posterior beliefs in feature space. As in the earlier work cited above, expectations are represented by inner products in respective state and observation feature spaces, and can be efficiently computed using the kernel trick. While the original Bellman operator has contractive and isotonic properties, i.e., the original value iteration is guaranteed to converge monotonically [Porta. et al., 2006], the resulting kernel Bellman operator does not. That said, these properties can be enforced following a simple correction, as proposed by [Grunewalder et al., 2012]. Approaches from the classical POMDP literature for initializing and enhancing efficiency can be applied, including ways to set initial values over the state feature space, and pruning methods for action edges.

Since kPOMDPs are nonparametric, the embeddings of the distributions used in the algorithm are learned from training samples. Note that in training, we must have access to samples from the hidden state, although for the test phase no such observations are necessary. This setting is reasonable in cases where hidden states are relatively costly to obtain: while it might be possible to observe them initially when learning the system dynamics, we would not have access to them during the value iteration phase, when learning an optimal policy.

Advantages of kPOMDPs include

- kPOMDPs can handle complex distributions over states and observations, since kPOMDPs are nonparametric.
- kPOMDPs can handle a wide variety of data types, as RKHS kernels have been defined on many domains: these include discrete, continuous, and structured data.
- kPOMDPs can be applied to high-dimensional POMDPs, since the computation scales with training sample size, and not with state and observation dimensionality. The convergence of the distribution embedding is at a rate independent of the dimension of the underlying space [Fukumizu et al., 2011, Gretton et al., 2012].

Our experiments confirm that the proposed kernel value iteration algorithm learns the correct policy on POMDP benchmark tasks.

This paper is organized as follows. We introduce the notations for POMDPs in the next section, and review recent kernel methods for probability distributions in Section 3. We present kernel POMDPs in Section 4, where Bellman equations in feature spaces (Subsection 4.1), the empirical expression and value iteration algorithms (Subsection 4.2) are shown. Experiments follow for an online planning algorithm.

## 2 POMDPs

A POMDP is specified by a tuple $\langle \mathcal{S}, \mathcal{A}, T, R, \mathcal{O}, Z \rangle$ where $\mathcal{S}$ is the set of states, $\mathcal{A}$ is the set of actions, $T : \mathcal{S} \times \mathcal{A} \times \mathcal{S} \to [0, \infty)$[1] is the transition function where $T(s, a, s') = \Pr(s'|s, a)$ represents the distribution of the next state $s'$ given an action $a$ in a state $s$, $R : \mathcal{S} \times \mathcal{A} \to \mathbb{R}$ is the reward function given an action $a$ in a state $s$, $\mathcal{O}$ is the set of observations, $Z : \mathcal{S} \times \mathcal{O} \to [0, \infty)$[2] is the observation function where $Z(s, o) = \Pr(o|s)$ represents the distribution of observation $o$ given a state $s$. We assume that $R$ is bounded.

An agent executes an action $a$ under the setting that the true state $s$ is not known, but partial information $o$ can be observed according to $Z(s, o)$. The agent then transitions to its next state $s'$ according to $T(s, a, s')$, receiving a reward $R(s, a)$ and making a new observation $o'$ according to $Z(s', o')$. The agent executes then its next action $a'$, and the process is repeated.

The goal of reinforcement learning in a POMDP is, given an initial belief $b_0$ for the initial state and a history of actions and observations $h_{t+1} = \{a_0, o_1, ..., a_t, o_{t+1}\}$, to find an optimal policy $\pi^*_{t+1} : (b_0, h_{t+1}) \mapsto a_{t+1}$ that maximizes the value function of the expected sum of discounted rewards with infinite horizon $\mathbb{E}[\sum_{t=0}^{\infty} \gamma^t R_t]$, where $\gamma \in (0, 1)$ is a discount factor.

In POMDPs, all the information $(b_0, h_{t+1})$ is condensed in a sufficient statistic of a belief distribution $b$ over the state set $\mathcal{S}$, and the optimal policy $\pi^*_{t+1}$ is reduced to $\pi^* : b \mapsto a$. Belief $b_{t+1} (t \geq 0)$ is updated

[1] In discrete case, $T : \mathcal{S} \times \mathcal{A} \times \mathcal{S} \to [0, 1]$.

[2] In discrete case, $Z : \mathcal{S} \times \mathcal{O} \to [0, 1]$.

according to Bayes' rule

$$b_{t+1}(s_{t+1}) = \frac{Z(s_{t+1}, o_{t+1})P(s_{t+1}|a_t; b_t)}{P(o_{t+1}|a_t; b_t)}, \tag{1}$$

where

$$P(s_{t+1}|a_t; b_t) = \mathbb{E}_{S_t \sim b_t}\left[T(S_t, a_t, s_{t+1})\right],$$
$$P(o_{t+1}|a_t; b_t) = \mathbb{E}_{S_{t+1} \sim P(\cdot|b_t, a_t)}\left[Z(o_{t+1}, S_{t+1})\right].$$

A variable with a prime indicates a value at the next timestep. $b^{a,o'}$ denotes the posterior belief for the next state $S'$ when the next observation $o'$ is observed after executing action $a$ in belief $b$.

The value function of belief $b$ following a fixed policy $\pi$ is given by

$$V^{\pi}(b) = \mathbb{E}\left[\sum_{t=0}^{\infty} \gamma^t \mathbb{E}_{S_t \sim b_t}\left[R(S_t, \pi(b_t))\right]\right], \tag{2}$$

and it is the fixed point of the Bellman equation

$$V^{\pi}(b) = Q^{\pi}(b, \pi(b)),$$
$$Q^{\pi}(b,a) = \mathbb{E}_{S \sim b}\left[R(S,a)\right] + \gamma \mathbb{E}_{O' \sim P(\cdot|b,a)}\left[V^{\pi}(b^{a,O'})\right] \tag{3}$$

where $Q^{\pi}(b,a)$ is the action value function, i.e., the value of executing action $a$ in belief $b$ while future actions $a', a'', \ldots$ are chosen according to policy $\pi$.

The optimal policy $\pi^*$ and its optimal value function $V^*(b)$ are given by the fixed point of the Bellman optimality equation

$$V^*(b) = \max_{a \in \mathcal{A}} Q^*(b,a),$$
$$Q^*(b,a) = \mathbb{E}_{S \sim b}\left[R(S,a)\right] + \gamma \mathbb{E}_{O' \sim p(\cdot|b,a)}\left[V^*(b^{a,O'})\right],$$
$$\pi^* = \arg\max_{a \in \mathcal{A}} Q^*(b,a). \tag{4}$$

$V^*(b)$ can be estimated by the value iteration algorithm $V_d = HV_{d-1} (d \geq 1)$, where $V_d$ is the $d$-step value function and $H$ is the Bellman operator

$$(HV)(b) = \max_{a \in \mathcal{A}}\left[\mathbb{E}_{S \sim b}\left[R(S,a)\right] + \gamma \mathbb{E}_{O' \sim p(\cdot|b,a)}\left[V(b^{a,O'})\right]\right]. \tag{5}$$

$H$ is known to be isotonic, contractive, and $V_d$ is guaranteed to be more accurate than $V_{d-1}$; that is, if $V_{d-1}$ satisfies $\varepsilon = \sup_b |V^*(b) - V_{d-1}(b)|$, then $V_d$ has an error bound of $|V^*(b) - \hat{V}_d(b)| \leq \gamma\varepsilon$ where $\gamma \in (0,1)$ [Ross et al., 2008].

The initial value $V_0(b)$ of belief $b$ used for value iteration can be defined using the initial $Q$-value $Q_0(s,a)$ over states and actions,

$$V_0(b) = \max_{a \in \mathcal{A}} \mathbb{E}_{S \sim b(\cdot)}\left[Q_0(S,a)\right]. \tag{6}$$

A simple choice for $Q_0(s,a)$ is the reward $Q_0(s,a) = R(s,a)$. Alternatively, a QMDP approximation $Q_0(s,a) = Q^{MDP}(s,a)$ may be used [Littman, 1995], where $Q^{MDP}(s,a)$ is given by running MDP value iteration, and approximating the POMDP by an MDP.

In kernel POMDPs, all the expectations appearing above are computed nonparametrically, without explicitly estimating the distributions $b(S)$, $P(S'|a;b)$, $P(O'|a;b)$, $b^{a,o'}(S')$ and transition and observation models, $T$ and $Z$. Instead, we obtain feature representations of the distributions (distribution embeddings) and the models (conditional embedding operators), as described in Section 3.

We end this section with a key to matching the probabilistic and reinforcement learning results presented in this section with their nonparametric, kernel-based counterparts in the following section. Bayes' rule (1) becomes the kernel Bayes' rule eq.(12), and its empirical counterpart leads to the updates (15)(16); the Bellman equations (3)(4) take the form of Claims 1-4, the Bellman operator (5) becomes the kernel Bellman operator (19), and the value initializations (6) lead to the initializations (20),(21).

# 3 Kernel Methods for Probabilities

In the present section, we provide an overview of distribution embeddings in reproducing kernel Hilbert spaces [Smola et al., 2007, Sriperumbudur et al., 2010, Gretton et al., 2012]. The embeddings are represented as mean (nonlinear) features of distributions, hence may also be referred to as *mean embeddings*. We also recall conditional embedding operators [Song et al., 2009, Song et al., 2010], and the Kernel Bayes' Rule (KBR) [Fukumizu et al., 2011]. Mean embeddings can be updated using conditional embedding operators; in particular, KBR allows us to obtain posterior embeddings given prior embeddings in feature spaces.

## 3.1 Embedding Distributions

Let $\mathcal{H}_{\mathcal{X}}$ be an RKHS associated with a bounded and measurable positive definite kernel $k_{\mathcal{X}} : \mathcal{X} \times \mathcal{X} \to \mathbb{R}$ over domain $(\mathcal{X}, \mathcal{B}_{\mathcal{X}})$, with $\langle \cdot, \cdot \rangle_{\mathcal{H}_{\mathcal{X}}}$ the corresponding inner product. The embedding of a distribution $P$ over $(\mathcal{X}, \mathcal{B}_{\mathcal{X}})$ in $\mathcal{H}_{\mathcal{X}}$ is given by the RKHS element $\mu_X = \mathbb{E}_{X \sim P}\left[k_{\mathcal{X}}(X, \cdot)\right] \in \mathcal{H}_{\mathcal{X}}$. $\mu_X$ coincides with the unique element satisfying $\langle \mu_X, f \rangle_{\mathcal{H}_{\mathcal{X}}} = \mathbb{E}_{X \sim P}\left[f(X)\right]$ for all $f \in \mathcal{H}_{\mathcal{X}}$, which means that the expectation of any function $f \in \mathcal{H}_{\mathcal{X}}$ can be computed as an inner product of the embedding $\mu_X$ and $f$ in $\mathcal{H}_{\mathcal{X}}$, without explicitly estimating the distribution $P$.

The element $\mu_X$ can be estimated by a linear combination of feature vectors on samples $(X_1, \ldots, X_n)$ over $\mathcal{X}$, i.e., $\hat{\mu}_X = \Upsilon\boldsymbol{\alpha}$ where $\boldsymbol{\alpha} = (\alpha_1, \ldots, \alpha_n)^\top \in \mathbb{R}^n$ and $\Upsilon = (k_\mathcal{X}(\cdot, X_1), \ldots, k_\mathcal{X}(\cdot, X_n))$. Using the empirical embedding $\hat{\mu}_X$, the expectation can be nonparametrically estimated as

$$\mathbb{E}_{X\sim P}\left[f(X)\right] \sim \langle \hat{\mu}_X, f\rangle_{\mathcal{H}_\mathcal{X}} = \boldsymbol{\alpha}^\top \mathbf{f}, \tag{7}$$

where $\mathbf{f} = (f(X_1), \ldots, f(X_n))^\top$.

We use characteristic kernels, e.g., Gaussian kernels and Laplacian kernels, which guarantee that each distribution maps to a unique embedding in the RKHS [Sriperumbudur et al., 2010, Fukumizu et al., 2011].

### 3.2 Conditional Embedding Operators & Kernel Bayes' Rule (KBR)

Let $\mathcal{H}_\mathcal{X}$ and $\mathcal{H}_\mathcal{Y}$ be RKHSs associated with $k_\mathcal{X}$ and $k_\mathcal{Y}$ over $(\mathcal{X}, \mathcal{B}_\mathcal{X})$ and $(\mathcal{Y}, \mathcal{B}_\mathcal{Y})$, respectively. Let $(X, Y)$ be a random variable taking values on $\mathcal{X} \times \mathcal{Y}$ with distribution $P$ and the density $p(x, y)$. The conditional density functions $\{p(Y|X = x)|x \in \mathcal{X}\}$ define a family of embeddings $\{\mu_{Y|x}\}$ in $\mathcal{H}_\mathcal{Y}$. According to [Song et al., 2009], a mapping from $k_\mathcal{X}(x, \cdot) \in \mathcal{H}_\mathcal{X}$ to $\mu_{Y|x} \in \mathcal{H}_\mathcal{Y}$ for all $x \in \mathcal{X}$ can be characterized by conditional embedding operator $\mathcal{U}_{Y|X} : \mathcal{H}_\mathcal{X} \to \mathcal{H}_\mathcal{Y}$,

$$\mu_{Y|x} = \mathcal{U}_{Y|X} k_\mathcal{X}(x, \cdot) = C_{YX} C_{XX}^{-1} k_\mathcal{X}(x, \cdot), \tag{8}$$

where $C_{YX}$ and $C_{XX}$ are uncentred covariance operators with respect to $P$,

$$\begin{aligned} C_{YX} &= \mathbb{E}_{(X,Y)\sim P}\left[k_\mathcal{Y}(Y, \cdot) \otimes k_\mathcal{X}(X, \cdot)\right], \\ C_{XX} &= \mathbb{E}_{(X,Y)\sim P}\left[k_\mathcal{X}(X, \cdot) \otimes k_\mathcal{X}(X, \cdot)\right], \end{aligned} \tag{9}$$

and we use the tensor product definition $(a \otimes b)c = a\langle b, c\rangle$. Feature representations of conditional distributions $\mu_{Y|x}$ are obtained by applying the operator $\mathcal{U}_{Y|X}$ to the feature map $k_\mathcal{X}(x, \cdot)$.

Since a posterior distribution is also written as a conditional distribution, the embedding of a posterior can be expressed as a conditional embedding operator [Fukumizu et al., 2011]. Let $\Pi$ be a prior distribution with density $\pi(x)$, and $(\bar{X}, \bar{Y})$ be a new random variable with distribution $Q$ corresponding to the density $q(x, y) = p(y|x)\pi(x)$. The embedding of a posterior $q(\bar{X}|\bar{Y} = y)$ given $y$ can be expressed by a corresponding conditional embedding operator $\mathcal{U}_{\bar{X}|\bar{Y}}$

$$\mu_{\bar{X}|y} = \mathcal{U}_{\bar{X}|\bar{Y}} k_\mathcal{Y}(y, \cdot) = C_{\bar{X}\bar{Y}} C_{\bar{Y}\bar{Y}}^{-1} k_\mathcal{Y}(y, \cdot), \tag{10}$$

where $C_{\bar{X}\bar{Y}}$ and $C_{\bar{Y}\bar{Y}}$ are covariance operators with respect to $Q$.

$$\begin{aligned} C_{\bar{X}\bar{Y}} &= \mathbb{E}_{(\bar{X},\bar{Y})\sim Q}\left[k_\mathcal{X}(\bar{X}, \cdot) \otimes k_\mathcal{Y}(\bar{Y}, \cdot)\right], \\ C_{\bar{Y}\bar{Y}} &= \mathbb{E}_{(\bar{X},\bar{Y})\sim Q}\left[k_\mathcal{Y}(\bar{Y}, \cdot) \otimes k_\mathcal{Y}(\bar{Y}, \cdot)\right]. \end{aligned} \tag{11}$$

See Appendix for further details and empirical estimates.

The inference underlying kPOMDPs is accomplished with the embedding operator $\mathcal{U}_{Y|X}$ and posterior embedding operator $\mathcal{U}_{\bar{X}|\bar{Y}}$.

## 4 Kernel POMDPs (kPOMDPs)

We now present our main results: we formulate POMDPs in feature spaces, and propose a kernelized value iteration algorithm. A key to the notation is given in Table 1.

### 4.1 Kernel Bellman Equations (KBEs)

To kernelize the Bellman equations, we introduce three RKHSs for state set $\mathcal{S}$, action set $\mathcal{A}$, and observation set $\mathcal{O}$. Let $\mathcal{H}_\mathcal{S}$, $\mathcal{H}_\mathcal{A}$, $\mathcal{H}_\mathcal{O}$ be RKHSs associated with bounded and measurable positive definite kernels $k_\mathcal{S}$, $k_\mathcal{A}$, $k_\mathcal{O}$ over $(\mathcal{S}, \mathcal{B}_\mathcal{S})$, $(\mathcal{A}, \mathcal{B}_\mathcal{A})$, $(\mathcal{O}, \mathcal{B}_\mathcal{O})$, respectively. $\langle\cdot,\cdot\rangle_{\mathcal{H}_\mathcal{S}}$, $\langle\cdot,\cdot\rangle_{\mathcal{H}_\mathcal{A}}$, $\langle\cdot,\cdot\rangle_{\mathcal{H}_\mathcal{O}}$ denote their respective inner products, and $\varphi(S)$, $\psi(A)$, $\phi(O)$ their feature vectors.

The relevant distributions $b(S)$, $P(S'|a; b)$, $P(O'|a; b)$, $b^{a,o'}(S')$ can be embedded in the corresponding RKHSs $\mathcal{H}_\mathcal{S}$, $\mathcal{H}_\mathcal{O}$ as follows:

$$\begin{aligned} \mu_S &= \mathbb{E}_{S\sim b(\cdot)}[\varphi(S)], \\ \mu_{S'|a;\mu_S} &= \mathbb{E}_{S'\sim p(\cdot|a;b)}[\varphi(S')], \\ \mu_{O'|a;\mu_S} &= \mathbb{E}_{O'\sim p(\cdot|a;b)}[\phi(O')], \\ \mu_{S'}^{a,o'} &= \mathbb{E}_{S'\sim b^{a,o'}(\cdot)}[\varphi(S')]. \end{aligned}$$

These embeddings are related via conditional embedding operators expressing transition models $T$, observation models $Z$, and posteriors.

Let $\mathcal{U}_{S'|S,A} : \mathcal{H}_\mathcal{S} \otimes \mathcal{H}_\mathcal{A} \to \mathcal{H}_\mathcal{S}$ be the conditional embedding operator for the transition model $T$, and $\mathcal{U}_{O|S} : \mathcal{H}_\mathcal{S} \to \mathcal{H}_\mathcal{O}$ be the operator for the observation model $Z$. Let $\mathcal{U}_{\bar{S}|\bar{O}}^{(a,\mu_S)} : \mathcal{H}_\mathcal{O} \to \mathcal{H}_\mathcal{S}$ be the posterior embedding operator corresponding to eq.(10), where we use the prior embedding $\mu_{S'|a;\mu_S}$ in the KBR.

The four embeddings above are related as

$$\begin{aligned} \mu_{S'|a;\mu_S} &= \mathcal{U}_{S'|S,A}\mu_S \otimes k_A(a, \cdot), \\ \mu_{O'|a;\mu_S} &= \mathcal{U}_{O|S}\mu_{S'|a;\mu_S}, \\ \mu_{S'}^{a,o'} &= \mathcal{U}_{\bar{S}|\bar{O}}^{(a,\mu_S)} k_O(o', \cdot). \end{aligned} \tag{12}$$

The sequence of mappings $\mu_S \mapsto \mu_{S'|a;\mu_S} \mapsto \mu_{O'|a;\mu_S} \mapsto \mu_{S'}^{a,o'}$ depends on the action $a$. A policy $\pi$ is defined to be a mapping from embeddings $\mu_S$ to actions. In evaluating policies, value functions $V(\cdot)$ are defined to be functions of embeddings $\mu_S$, and thus (indirectly) of the actions $a$ and observations $o$.

Table 1: Notation

| | STATE | ACTION | OBSERVATION | (STATE, ACTION) |
|---|---|---|---|---|
| DOMAIN | $\mathcal{S}$ | $\mathcal{A}$ | $\mathcal{O}$ | $\mathcal{S}\times\mathcal{A}$ |
| R.V. | $S$ | $A$ | $O$ | $(S,A)$ |
| INSTANCE | $s$ | $a$ | $o$ | $(s,a)$ |
| KERNEL | $k_{\mathcal{S}}(\cdot,\cdot)$ | $k_{\mathcal{A}}(\cdot,\cdot)$ | $k_{\mathcal{O}}(\cdot,\cdot)$ | $k_{\mathcal{S}\times\mathcal{A}}(\cdot,\cdot)=k_{\mathcal{S}}(\cdot,\cdot)\otimes k_{\mathcal{A}}(\cdot,\cdot)$ |
| RKHS | $\mathcal{H}_{\mathcal{S}}$ | $\mathcal{H}_{\mathcal{A}}$ | $\mathcal{H}_{\mathcal{O}}$ | $\mathcal{H}_{\mathcal{S}\times\mathcal{A}}=\mathcal{H}_{\mathcal{S}}\otimes\mathcal{H}_{\mathcal{A}}$ |
| FEATURE MAP | $\varphi(s)=k_{\mathcal{S}}(s,\cdot)$ | $\psi(a)=k_{\mathcal{A}}(a,\cdot)$ | $\phi(o)=k_{\mathcal{O}}(o,\cdot)$ | $\vartheta(s,a)=k_{\mathcal{S}\times\mathcal{A}}((s,a),\cdot)$ |
| FINITE SAMPLE SET | $\mathcal{S}_0$ | $\mathcal{A}_0$ | $\mathcal{O}_0$ | $(\mathcal{S}_0,\mathcal{A}_0)$ |
| FEATURE MATRIX | $\Upsilon$ | $\Psi$ | $\Phi$ | $\Theta$ |
| GRAM MATRIX | $G_S=\Upsilon^\top\Upsilon$ | $G_A=\Psi^\top\Psi$ | $G_O=\Phi^\top\Phi$ | $G_{(S,A)}=G_S\odot G_A$ |
| FEATURE COLUMN | $\mathbf{k}_S(s)=\Upsilon^\top\varphi(s)$ | $\mathbf{k}_A(a)=\Psi^\top\psi(a)$ | $\mathbf{k}_O(o)=\Phi^\top\phi(o)$ | $\mathbf{k}_{(S,A)}(s,a)=\Theta^\top\vartheta(a,o)$ |

Using value functions over embeddings, the expected immediate rewards and expected values of posterior beliefs in the Bellman equations (Section 2) can be computed as inner products in RKHSs,

$$\mathbb{E}_{S\sim b}\left[R(S,a)\right]=\langle\mu_S,R(\cdot,a)\rangle_{\mathcal{H}_{\mathcal{S}}},$$
$$\mathbb{E}_{O'\sim P(\cdot|b,a)}\left[V^\pi(b^{a,O'})\right]=\left\langle\mu_{O'|a;\mu_S},V^\pi\left(\mu_{S'}^{a,(\cdot)}\right)\right\rangle_{\mathcal{H}_{\mathcal{O}}},$$
$$\mathbb{E}_{O'\sim P(\cdot|b,a)}\left[V^*(b^{a,O'})\right]=\left\langle\mu_{O'|a;\mu_S},V^*\left(\mu_{S'}^{a,(\cdot)}\right)\right\rangle_{\mathcal{H}_{\mathcal{O}}},$$

assuming $R(\cdot,a)\in\mathcal{H}_{\mathcal{S}}$ and $V^\pi\left(\mu_{S'}^{a,(\cdot)}\right),V^*\left(\mu_{S'}^{a,(\cdot)}\right)\in\mathcal{H}_{\mathcal{O}}$.

We are now ready to introduce the kernel Bellman equation as an operation on the embeddings $\mu_S$. Let $\mathcal{P}_{\mathcal{S}}$ be the set of beliefs and $\mathcal{I}_{\mathcal{S}}$ be the set of embeddings of $\mathcal{P}_{\mathcal{S}}$ in $\mathcal{H}_{\mathcal{S}}$.

**Claim 1.** *Let $R(\cdot,a)\in\mathcal{H}_{\mathcal{S}}$ and $V^\pi\left(\mu_{S'}^{a,(\cdot)}\right)\in\mathcal{H}_{\mathcal{O}}$ for all $a\in\mathcal{A}$ and $\mu_S\in\mathcal{I}_{\mathcal{S}}$. The kernel Bellman equations on $\mathcal{H}_{\mathcal{S}}$ are*

$$V^\pi(\mu_S)=Q^\pi\left(\mu_S,\pi\left(\mu_S\right)\right),$$
$$Q^\pi(\mu_S,a)=\langle\mu_S,R\left(\cdot,a\right)\rangle_{\mathcal{H}_{\mathcal{S}}}+\gamma\left\langle\mu_{O'|a;\mu_S},V^\pi\left(\mu_{S'}^{a,(\cdot)}\right)\right\rangle_{\mathcal{H}_{\mathcal{O}}}.$$

The solution of these equations following a fixed policy $\pi$ on $\mathcal{H}_{\mathcal{S}}$ yields a value function $V^\pi(\cdot)$. The kernel Bellman optimality equations take similar form.

**Claim 2.** *Let $R(\cdot,a)\in\mathcal{H}_{\mathcal{S}}$ and $V^*\left(\mu_{S'}^{a,(\cdot)}\right)\in\mathcal{H}_{\mathcal{O}}$ for all $a\in\mathcal{A}$ and $\mu_S\in\mathcal{I}_{\mathcal{S}}$. The kernel Bellman optimality equations on RKHS $\mathcal{H}_{\mathcal{S}}$ are*

$$V^*(\mu_S)=\max_{a\in\mathcal{A}}Q^*(\mu_S,a),$$
$$Q^*(\mu_S,a)=\langle\mu_S,R\left(\cdot,a\right)\rangle_{\mathcal{H}_{\mathcal{S}}}+\gamma\left\langle\mu_{O'|a;\mu_S},V^*\left(\mu_{S'}^{a,(\cdot)}\right)\right\rangle_{\mathcal{H}_{\mathcal{O}}},$$
$$\pi^*(\mu_S)=\arg\max_{a\in\mathcal{A}}Q^*\left(\mu_S,a\right). \tag{13}$$

The solution of these equations is the optimal value function $V^*(\cdot)$ with corresponding optimal policy $\pi^*(\cdot)$ on $\mathcal{H}_{\mathcal{S}}$. We use the following tuple for solving a POMDP in feature space:

$$\left\langle\mathcal{H}_{\mathcal{S}},\mathcal{H}_{\mathcal{A}},\mathcal{U}_{S|S,A},R,\mathcal{H}_{\mathcal{O}},\mathcal{U}_{O|S},\mathcal{U}_{\bar{S}|\bar{O}}^{(\mathcal{A},\mathcal{I}_{\mathcal{S}})}\right\rangle,$$

where $\mathcal{U}_{\bar{S}|\bar{O}}^{(\mathcal{A},\mathcal{I}_{\mathcal{S}})}=\{\mathcal{U}_{\bar{S}|\bar{O}}^{(a,\mu_S)}|a\in\mathcal{A},\mu_S\in\mathcal{I}_{\mathcal{S}}\}$.

### 4.2 Empirical Expression

We next provide a finite sample version of the kernel POMDP. Suppose that $D_n=\{(\tilde{s}_i,\tilde{o}_i),\tilde{a}_i,\tilde{R}_i,(\tilde{s}'_i,\tilde{o}'_i)\}_{i=1}^n$ are $n$ training samples according to a POMDP $\langle\mathcal{S},\mathcal{A},T,R,\mathcal{O},Z\rangle$. Note that state samples $\{(\tilde{s}_i,\tilde{s}'_i)\}$ are included, and will used for estimating belief embeddings $\mu_S$, $\mu_{S'}^{a,o'}$. We assume that such samples from the true state are available for training, but not during the test phase.

Let $\mathcal{S}_0$, $\mathcal{O}_0$, $\mathcal{A}_0$, $\mathcal{S}'_0$, $\mathcal{O}'_0$ be finite sets of states, observations, and actions taken from the training samples $D_n$. The associated feature vectors are

$$\Upsilon=\left(\varphi\left(\tilde{s}_1\right),\ldots,\varphi\left(\tilde{s}_n\right)\right),\Upsilon'=\left(\varphi\left(\tilde{s}'_1\right),\ldots,\varphi\left(\tilde{s}'_n\right)\right),$$
$$\Phi=\left(\phi\left(\tilde{o}_1\right),\ldots,\phi\left(\tilde{o}_n\right)\right),\Phi'=\left(\phi\left(\tilde{o}'_1\right),\ldots,\phi\left(\tilde{o}'_n\right)\right),$$
$$\Psi=\left(\psi\left(\tilde{a}_1\right),\ldots,\psi\left(\tilde{a}_n\right)\right),\Theta=\left(\vartheta\left(\tilde{s}_1,\tilde{a}_1\right),\ldots,\vartheta\left(\tilde{s}_n,\tilde{a}_n\right)\right).$$

We build Gram matrices from the feature matrices $\Upsilon$, $\Psi$, $\Phi$ as $G_S=\Upsilon^\top\Upsilon$, $G_A=\Psi^\top\Psi$, $G_O=\Phi^\top\Phi$, $G_{(S,A)}=G_S\odot G_A$, where $\odot$ denotes the Hadamard (element-wise) product. The embeddings $\mu_S$, $\mu_{O'|a;\mu_S}$, $\mu_{S'}^{a,o'}$ take respective forms

$$\begin{aligned}\hat{\mu}_S&=\Upsilon\boldsymbol{\alpha},\\ \hat{\mu}_{O'|a;\mu_S}&=\Phi\boldsymbol{\beta}'_{a;\boldsymbol{\alpha}},\\ \hat{\mu}_{S'}^{a,o'}&=\Upsilon\boldsymbol{\alpha}'_{a,o'}.\end{aligned} \tag{14}$$

The update rules for the weight vectors $\boldsymbol{\alpha}\mapsto\boldsymbol{\beta'}_{a;\boldsymbol{\alpha}}\mapsto\boldsymbol{\alpha}'_{a,o'}$ corresponding to $\mu_S\mapsto\mu_{O'|a;\mu_S}\mapsto\mu_{S'}^{a,o'}$ in eq.(12) are given as follows:

- The update $\boldsymbol{\alpha}\mapsto\boldsymbol{\beta'}_{a;\boldsymbol{\alpha}}$ uses the empirical estimates of conditional embedding operators $\mathcal{U}_{S'|S,A}$, $\mathcal{U}_{O|S}$, which results in a linear transformation $\boldsymbol{\beta}'_{a;\boldsymbol{\alpha}} = L_{O|S,a}\boldsymbol{\alpha}$ for all $a \in \mathcal{A}$ by the $n \times n$ matrix $L_{O|S,a}$:

$$(G_S+\varepsilon_S nI_n)^{-1}G_{SS'}\left(G_{(S,A)}+\varepsilon_{(S,A)}nI_n\right)^{-1}G_{(S,A)(S,a)} \tag{15}$$

with $G_{SS'} := \Upsilon^\top \Upsilon'$, $G_{(S,A)(S,a)} := D\left(\mathbf{k}_A(a)\right)G_S$, and $\mathbf{k}_A(a) = \Psi^\top \psi(a)$.

- The update $\boldsymbol{\beta}'_{a;\boldsymbol{\alpha}} \mapsto \boldsymbol{\alpha}'_{a,o'}$ is based on the kernel Bayes' rule. The Gram matrix expression, eq.(25), results in $\boldsymbol{\alpha}'_{a,o'} = R_{S|O}(\hat{\boldsymbol{\beta}}'_{a;\boldsymbol{\alpha}})\mathbf{k}_O(o')$ using a non-negative vector $\hat{\boldsymbol{\beta}}'_{a;\boldsymbol{\alpha}}$ and an $n \times n$ matrix $R_{S|O}(\hat{\boldsymbol{\beta}}'_{a;\boldsymbol{\alpha}})$:

$$\left(D(\hat{\boldsymbol{\beta}}'_{a;\boldsymbol{\alpha}})G_O + \epsilon nI_n\right)^{-1}D(\hat{\boldsymbol{\beta}}'_{a;\boldsymbol{\alpha}}). \tag{16}$$

Given samples $D_n$, the embeddings $\hat{\mu}_S, \hat{\mu}_{O'|a;\mu_S}, \hat{\mu}_{S'}^{a,o'}$ are identified with $n$-dimensional vectors $\boldsymbol{\alpha}, \boldsymbol{\beta'}_{a;\boldsymbol{\alpha}}, \boldsymbol{\alpha}'_{a;\boldsymbol{\alpha}} \in \mathbb{R}^n$, respectively, and the Bellman equations (Claims 1, 2) can be represented in terms of these weight vectors. Therefore, the belief and predictive distributions are represented by $n$-dimensional vectors for any sets $\mathcal{S}$ and $\mathcal{O}$.

**Claim 3.** *Given samples $D_n$, the kernel Bellman equation (Claim 1) has the empirical expression*

$$\begin{aligned}\hat{V}^\pi(\boldsymbol{\alpha}) &= \hat{Q}^\pi(\boldsymbol{\alpha}, \pi(\boldsymbol{\alpha})),\\ \hat{Q}^\pi(\boldsymbol{\alpha}, a) &= \boldsymbol{\alpha}^\top \boldsymbol{R}_a + \gamma{\boldsymbol{\beta}'}^\top_{a;\boldsymbol{\alpha}}\hat{\mathbf{V}}^\pi\left(\boldsymbol{\alpha}'_{a,\mathcal{O}_0}\right),\end{aligned} \tag{17}$$

*where $\mathbf{R}_a = (R(\tilde{s}_1, a), \ldots, R(\tilde{s}_n, a))^\top \in \mathbb{R}^n$ is the reward vector on samples $\mathcal{S}_0$ for action $a$ and $\hat{\mathbf{V}}^\pi\left(\boldsymbol{\alpha}'_{a,\mathcal{O}_0}\right) = \left(\hat{V}^\pi\left(\boldsymbol{\alpha}'_{a,\tilde{o}_1}\right), \ldots, \hat{V}^\pi\left(\boldsymbol{\alpha}'_{a,\tilde{o}_n}\right)\right)^\top \in \mathbb{R}^n$ is the posterior belief value vector on samples $\mathcal{O}_0$ given action $a$.*

**Claim 4.** *Given samples $D_n$, the kernel Bellman optimality equation (Claim 2) has the expression*

$$\begin{aligned}\hat{V}^*(\boldsymbol{\alpha}) &= \max_{a\in\mathcal{A}} \hat{Q}^*(\boldsymbol{\alpha}, a),\\ \hat{Q}^*(\boldsymbol{\alpha}, a) &= \boldsymbol{\alpha}^\top \boldsymbol{R}_a + \gamma{\boldsymbol{\beta}'}^\top_{a;\boldsymbol{\alpha}}\mathbf{V}^*\left(\boldsymbol{\alpha}'_{a,\mathcal{O}_0}\right),\\ \hat{\pi}^*(\boldsymbol{\alpha}) &= \arg\max_{a\in\mathcal{A}} \hat{Q}^*(\boldsymbol{\alpha}, a),\end{aligned} \tag{18}$$

*where $\mathbf{R}_a = (R(\tilde{s}_1, a), \ldots, R(\tilde{s}_n, a))^\top \in \mathbb{R}^n$ and $\hat{\mathbf{V}}^*\left(\boldsymbol{\alpha}'_{a,\mathcal{O}_0}\right) = \left(\hat{V}^*\left(\boldsymbol{\alpha}'_{a,\tilde{o}_1}\right), \ldots, \hat{V}^*\left(\boldsymbol{\alpha}'_{a,\tilde{o}_n}\right)\right)^\top \in \mathbb{R}^n$.*

Figure 1 illustrates the planning forward search using the Bellman equations. Even when the assumptions in Claims 1, 2 do not hold, we use Claims 3, 4 as an approximation, and Claim 4 for value iteration. Let $\hat{H}_n$ be the kernel Bellman operator,

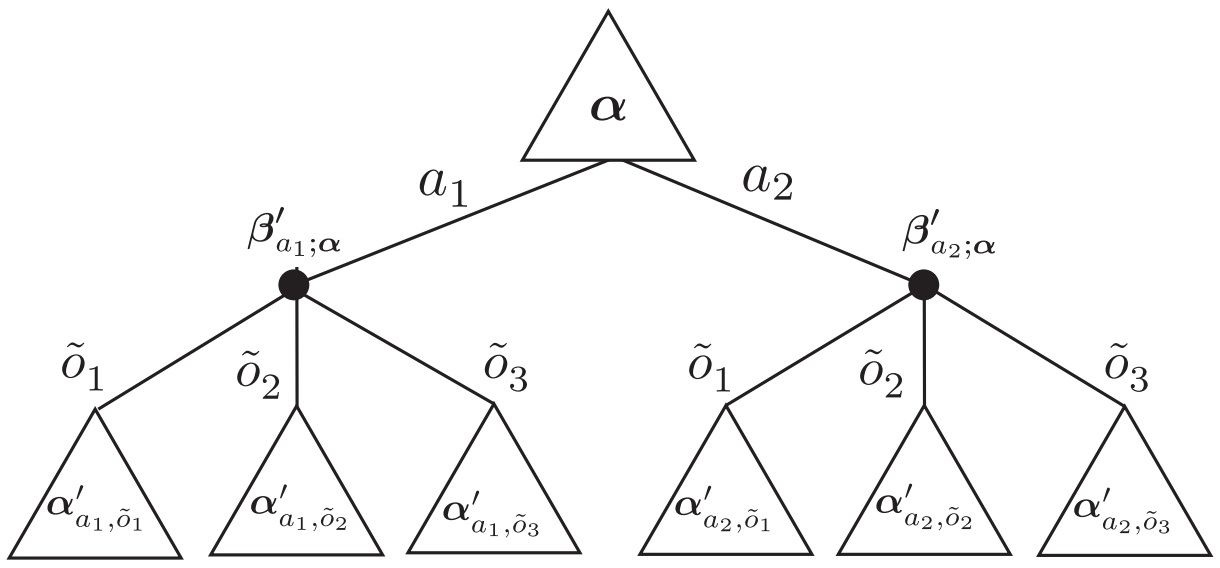

Figure 1: A search tree using kernel Bellman equations. Beliefs are represented by $n$-dimensional weight vectors $\boldsymbol{\alpha}$ on samples. The expected immediate reward (first term) is given by the linear combination $\boldsymbol{\alpha}^\top \boldsymbol{R}_a$ on samples $\mathcal{S}_0$ and associated with the action link $a$. The discounted expected value for next beliefs (second term) is given by the linear combination $\gamma{\boldsymbol{\beta}'}^\top_{a;\boldsymbol{\alpha}}\hat{\mathbf{V}}\left(\boldsymbol{\alpha}'_{a,\mathcal{O}_0}\right)$ on samples $\mathcal{O}_0$ and associated with the observation links $\mathcal{O}_0 = \{\tilde{o}_1, \ldots, \tilde{o}_n\}$. Observation links are expanded with respect to finite set $\mathcal{O}_0$ instead of $\mathcal{O}$. The kernel Bayes' rule is applied to each pair $(a, \tilde{o})$. Values are back-propagated bottom to top starting at initial values $V_0(\cdot)$ in the kernel value iteration algorithm.

$$(\hat{H}_n V)(\boldsymbol{\alpha}) = \max_{a\in\mathcal{A}}\left[\boldsymbol{\alpha}^\top \boldsymbol{R}_a + \gamma{\boldsymbol{\beta}'}^\top_{a;\boldsymbol{\alpha}}\mathbf{V}\left(\boldsymbol{\alpha}'_{a,\mathcal{O}_0}\right)\right]. \tag{19}$$

We run value iteration $\hat{V}_d = \hat{H}_n\hat{V}_{d-1} (d \geq 1)$ with an initial value function $V_0(\cdot)$ on weights. The detailed algorithm is shown in Algorithm 1. The computational complexity is given in Subsection 4.3. We use the same value initializations as in distributional POMDPs (Section 2). If $Q_0(\cdot, a) \in \mathcal{H}_\mathcal{S}$ for all $a \in \mathcal{A}$, the initial values $V_0(\cdot)$ over embeddings can be set as

$$V_0(\mu_S) = \max_{a\in\mathcal{A}} \langle \mu_S, Q_0(\cdot, a)\rangle_{\mathcal{H}_\mathcal{S}}, \tag{20}$$

which leads to the empirical expression

$$V_0(\boldsymbol{\alpha}) = \max_{a\in\mathcal{A}} \boldsymbol{\alpha}^\top \mathbf{Q}_a^0, \tag{21}$$

where $\mathbf{Q}_a^0 = \left(Q_0(\tilde{s}_1, a), \ldots, Q_0(\tilde{s}_n, a)\right)^\top \in \mathbb{R}^n$ is an initial action value vector on samples $\mathcal{S}_0$. The reward function $Q_0(s, a) = R(s, a)$ or the QMDP value $Q_0(s, a) = Q^{MDP}(s, a)$ can be used for initialization.

Since $\boldsymbol{\alpha}, \boldsymbol{\beta'}_{a;\boldsymbol{\alpha}}, \boldsymbol{\alpha}'_{a;\boldsymbol{\alpha}}$ do not always give nonnegative vectors, the Bellman operator $\hat{H}_n$ is not guaranteed to be isotonic and contractive, although empirically, $\hat{H}_n$ can be used for the value iteration algorithm. $\hat{H}_n$ is corrected to have the isotonic and contractive properties by approximating the weight vectors as probability

**Algorithm 1** Kernel Value Iteration ($\boldsymbol{\alpha}$, $d$)

**Input:** belief weights $\boldsymbol{\alpha}$, tree depth $d$
**Output:** Value $V_d(\boldsymbol{\alpha})$, Policy $\pi_d(\boldsymbol{\alpha})$
**if** $d = 0$ **then**
  Set: $V_d(\boldsymbol{\alpha}) = \max_{a \in \mathcal{A}} Q_0(\boldsymbol{\alpha}, a)$,
    $\pi_d(\boldsymbol{\alpha}) = \arg\max_{a \in \mathcal{A}} Q_0(\boldsymbol{\alpha}, a)$
**else**
  **for** all $a \in \mathcal{A}$ **do**
    Compute: $\boldsymbol{\beta}'_{a;\boldsymbol{\alpha}} = L_{O|S,a}\boldsymbol{\alpha}$
    **for** all $\tilde{o}_i \in \mathcal{O}_0$ **do**
      **if** $\left(\boldsymbol{\beta}'_{a;\boldsymbol{\alpha}}\right)_i \neq 0$ **then**
        Compute posterior:
        $\boldsymbol{\alpha}'_{a,\tilde{o}_i} = R_{S|O}(\boldsymbol{\beta}'_{a;\boldsymbol{\alpha}})\mathbf{k}_O(\tilde{o}_i)$
        Set:
        $V(\boldsymbol{\alpha}'_{a,\tilde{o}_i}) \leftarrow$ Kernel Val. Iter.$(\boldsymbol{\alpha}'_{a,\tilde{o}_i}, d{-}1)$
      **end if**
    **end for**
  **end for**
  Set: $V_d(\boldsymbol{\alpha}) = \max_{a \in \mathcal{A}} Q_d(\boldsymbol{\alpha}, a)$,
    $\pi_d(\boldsymbol{\alpha}) = \arg\max_{a \in \mathcal{A}} Q_d(\boldsymbol{\alpha}, a)$
**end if**

vectors [Grunewalder et al., 2012]. Let $\hat{\boldsymbol{\alpha}}, \hat{\boldsymbol{\beta}}'_{a;\boldsymbol{\alpha}}, \hat{\boldsymbol{\alpha}}'_{a;\boldsymbol{\alpha}}$ be probability vectors derived from $\boldsymbol{\alpha}, \boldsymbol{\beta}'_{a;\boldsymbol{\alpha}}, \boldsymbol{\alpha}'_{a;\boldsymbol{\alpha}}$ according to $\hat{w}_i = \frac{\max\{w_i, 0\}}{\sum_{i=1}^n \max\{w_i, 0\}}$ for weight vectors $\boldsymbol{w}$. The corresponding kernel Bellman operator $\hat{H}_n^+$ using the probability vectors $\hat{\boldsymbol{\alpha}}, \hat{\boldsymbol{\beta}}'_{a;\boldsymbol{\alpha}}$ is guaranteed to be isotonic and contractive. The proof is given in the Supplementary material.

Given samples $D_n$, we use the following tuple for the kernel value iteration:

$$\left\langle (\mathcal{H}_{\mathcal{S}}, \mathbf{k}_S), (\mathcal{H}_{\mathcal{A}}, \mathbf{k}_A), R, (\mathcal{H}_{\mathcal{O}}, \mathbf{k}_O), L_{O|S,\mathcal{A}}, R_{S|O}(\cdot) \right\rangle,$$

where $L_{O|S,\mathcal{A}} = \{L_{O|S,a} | a \in \mathcal{A}\}$.

### 4.3 Computational Complexity

The update rule $\boldsymbol{\alpha} \mapsto \boldsymbol{\beta'}_{a;\boldsymbol{\alpha}}$ has complexity $O(n^2)$ with respect to the number of samples $n$ for an action $a$, where the matrix $L_{S|O,a}$ is computed only once in the training phase. The update rule $\boldsymbol{\beta}'_{a;\boldsymbol{\alpha}} \mapsto \boldsymbol{\alpha}'_{a,o'}$ has complexity $O(n^3)$ for an observation $o'$, due to the inversion of an $n \times n$ matrix (eq. 16). In total, the computation of the posterior weights $\boldsymbol{\alpha}'_{a,o'}$ for a pair $(a, o')$ costs $O(n^3)$, compared with $O(|\mathcal{S}|^2)$ for Bayes' rule in eq.(1). Kernel value iteration to depth $d$ costs $O(n^3(n|\mathcal{A}|)^d)$, whereas classical value iteration costs $O(|\mathcal{S}|^2(|\mathcal{O}||\mathcal{A}|)^d)$. The complexity $O(n^3)$ for computing $\boldsymbol{\alpha}'_{a,o'}$ can be reduced to $O(nr^2)$ via a low rank approximations of the $n \times n$ Gram matrices, where $r$ is the rank.

**Algorithm 2** Online Planning with Finite Horizon $d$

Set: $T$, $t = 0$
Get: Initial observation $o$
Compute: Initial belief $\boldsymbol{\alpha} = (G_O + n\epsilon_O I_n)^{-1}\mathbf{k}_O(o)$
**repeat**
  Planning: $a \leftarrow$ Kernel Value Iteration ($\boldsymbol{\alpha}$, $d$)
  Get: reward and new observation $(R, o')$
  Compute next belief: $\boldsymbol{\alpha} = R_{S|O}(\boldsymbol{\beta}'_{a;\boldsymbol{\alpha}})\mathbf{k}_O(o')$
  $t = t + 1$
**until** $t > T$

## 5 Experiments

We implemented an online planning algorithm in POMDPs using the kernel Bellman equations (Algorithm 2). An example of kPOMDP dynamics is shown in Figure 3. An agent computes the initial belief weights $\boldsymbol{\alpha} \in \mathbb{R}^n$ by $\boldsymbol{\alpha} = (G_O + n\epsilon_O I_n)^{-1}\mathbf{k}_O(o)$ with an initial observation $o$, which corresponds to a belief estimate without a prior. The agent makes a decision $a$ by kernel value iteration to finite horizon $d$ under the current belief weights $\boldsymbol{\alpha}$. The agent then updates the belief weights using a function $R_{S|O}(\cdot)$ and predictive weights $\boldsymbol{\beta}'_{a;\boldsymbol{\alpha}}$ when obtaining a reward $R$ and a next observation $o'$. When the prediction fails (e.g., $D(\boldsymbol{\beta}'_{a;\boldsymbol{\alpha}})\mathbf{k}_O(o') = \mathbf{0}$, or in discrete cases $\boldsymbol{\beta}'_{a;\boldsymbol{\alpha}}(o') = 0$) in the case of small training samples, we reset current belief weights and estimated initial belief weights. To make the online planning algorithms more efficient, we computed the inverses of the $n \times n$ matrices in our algorithm by combining low rank approximations based on the incomplete Cholesky factorization and the Woodbury identity [Fine and Scheinberg, 2001]

Figure 2 shows some results on benchmarks [Littman, 1995], where the state, action, and observation sets are finite. Results are $d = 2$, $d = 1$, $d = 1$ for $10 \times 10$ Grid World, Network, Hallway, respectively. The exact policy is computed by an agent having complete knowledge about the POMDP environment. The histogram policy computed by running a classical value iteration algorithm, where transition and observation models are estimated by histograms using the same samples as our algorithm. Since the histogram policy requires samples over all the combinations of states and actions to estimate transition models, we introduced prior samples drawn from a uniform prior to test kPOMDP and Histograms under exactly the same conditions. Note that KBR does not need prior samples. We used QMDP initial values and pruned action links based on the QMDP values [Ross et al., 2008]. An action link was pruned if its QMDP value was lower than the current estimated value. The KBR-controller learned the exact policy as the number of training samples was increased in the $10 \times 10$ grid world and network problems. In our limited experiments, though KBR and histogram methods sometimes showed different results depending on training data, they gave similar results on average.

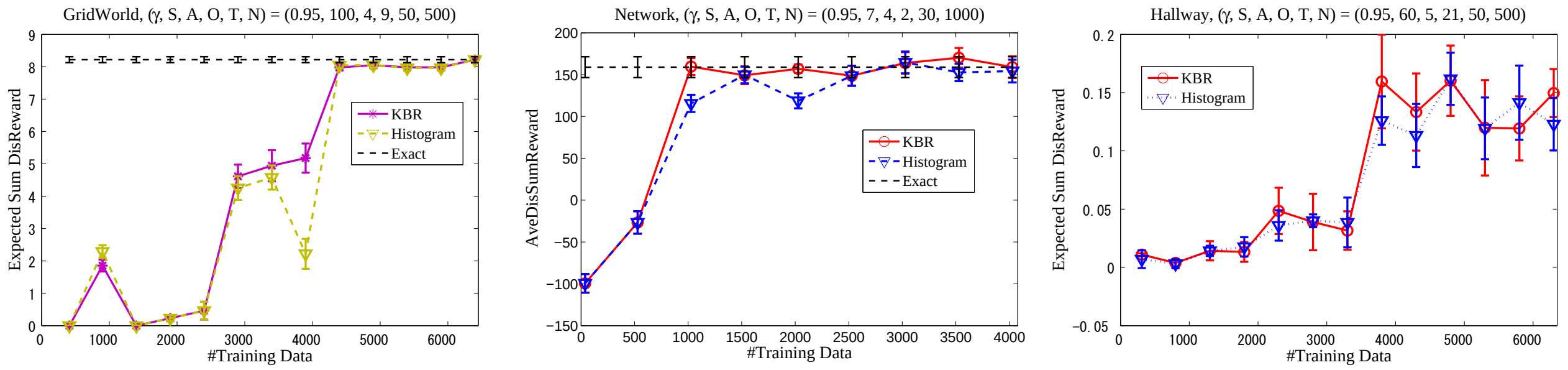

Figure 2: kPOMDPs+KBR controller. The left, middle, and right figures are $10 \times 10$ Gridworld, Network, and Hallway problems, respectively. Averaged discounted sum of rewards an agent got in test experiments are plotted against the number of training samples. We ran $N$ experiments, where one experiment consists of $T$ steps. The parameters are accompanied with the titles, ($\gamma$, S, A, O, T, N) = ($\gamma$, $|\mathcal{S}|$, $|\mathcal{A}|$, $|\mathcal{O}|$, $T$, $N$). Training samples are collected by uniform random actions. The leftmost plot for each figure is the result of uniform prior samples where no information about the environment. $10 \times 10$ grid world problem is similar to the $4 \times 4$ grid world, where state size is 100, reward is delivered by any action at a goal state with value 1, otherwise 0, and observations are 9 wall patterns about 4 nearest neighbors.

We also implemented a simulator of a swing-up cart-balancing system. The system consists of a cart with mass 8 kg running on a 2 m track and a freely swinging pendulum with mass 2 kg attached to the cart with a 50 cm rod. The state of the system is the angle and the angular velocity of the pendulum $(\theta, \dot{\theta})$, however the agent observes only the angle. The agent may apply a horizontal force or action to the cart, chosen from $a \in \{-250, -150, -50, 0, 50, 150, 250\}$N. The dynamics of the system are nonlinear. The states are continuous, but time is discretized in steps of 0.1 s. The objective is to balance the pendulum in the inverted position. Training samples are collected by applying uniform random actions from uniformly random states over $\theta = [-\pi/3, \pi/3]$, $\dot{\theta} = [3, 3]$. The reward function is given by $R(\theta, \dot{\theta}) = \exp(-\theta^2/2\sigma_1^2 - \dot{\theta}^2/10\sigma_2^2)$, where $\sigma_1^2$ and $\sigma_2^2$ are the variances of uniform distributions over $\theta = [-\pi/3, \pi/3]$, $\dot{\theta} = [3, 3]$, respectively.

Figure 3 shows a visualization of the kPOMDP dynamics and the inverted pendulum results (see caption details). The rightmost figure plots the averaged result of the earned rewards (height=cos($\theta$)) of the learned policies as a function of training samples. The episode length was 10 sec, i.e., the maximum of the total rewards is 100. The result was averaged over 20 experiments, the planning depth was $d = 1$, and the initial value function was the reward function $R(\theta, \dot{\theta})$. In kPOMDP, we used Gaussian kernels $G(\cdot, \cdot)$ for states and observations $k_{\mathcal{S}}((\theta_1, \dot{\theta}_1), (\theta_2, \dot{\theta}_2)) = G(\theta_1, \theta_2)G(\dot{\theta}_1, \dot{\theta}_2)$, $k_{\mathcal{O}}(\theta_1, \theta_2) = G(\theta_1, \theta_2)$, where the kernel parameters $\sigma$ were $\sigma = \text{MedDist}/30$ for $\theta$ and $\sigma = \text{MedDist}/10$ for $\dot{\theta}$, where MedDist is the median inter-sample distance. The kernel for actions was the identity. We compared kPOMDP to histogram policies where the environment is discretized. Histogram($M$) indicates $M \times M$ discretized states over $[-\pi/3, \pi/3] \times [3, 3]$. kPOMDP almost reached the maximum value 100 and showed better results than histogram policies.

# 6 Summary

We have introduced POMDPs in feature spaces, where beliefs over states are represented as distribution embeddings in feature spaces and updated via the kernel Bayes' rule. The Bellman equations, value functions, and policies are all expressed as functions of this feature representation. We further proposed a policy learning strategy by value iteration in the kernel framework, where the isotonic and contraction properties of the kernel Bellman operator are enforced by a simple correction. Value initialization and action edge pruning can be implemented for kernel POMDPs, following the approach of distributional POMDPs such as QMDP. Experiments confirm that the controller learned in feature space converges to the optimum policy. Our approach serves as a first step towards more powerful kernel-based algorithms for POMDPs.

# A Appendix: Kernel Bayes' Rule

A variant of the kernel Bayes' rule algorithm we have used in this paper is described. [Fukumizu et al., 2011] propose a squared regularization form for the empirical

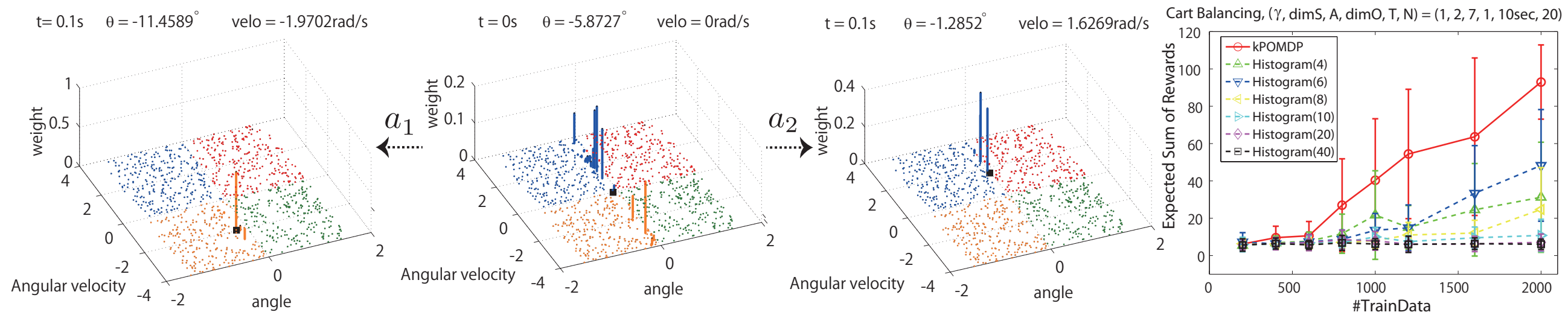

Figure 3: Inverted pendulum results with an example of kPOMDP dynamics. In the three 3D plots, all the points indicate training samples on state set $\mathcal{S} = (\theta, \dot{\theta})$ and $z$ axis indicates the magnitude of weights on their samples (after normalization), i.e., belief embedding weights $\hat{\boldsymbol{\alpha}}$ identifying belief embedding $\hat{\mu}_S$. The true state of the system is also marked by the black point in each 3D plot. The four colors indicate different combinations of signs of states (for instance, blue denotes positive angular velocity and negative angle). The middle figure in the 3D plots shows the initial belief estimate given an initial observation $o = \theta$, estimated by $\boldsymbol{\alpha} = (G_O + n\epsilon_O I_n)^{-1}\mathbf{k}_O(o)$ in Algorithm 2. Since $\dot{\theta}$ is uncertain at the initial point, positive weights spread in the direction of $\dot{\theta}$ axis. The left and right figures in the 3D plots correspond to the updated weights $\boldsymbol{\alpha}'_{a,o'}$ of belief embeddings depending on the executed actions $a_1$ (positive) and $a_2$ (negative) after observing a new observation $o' = \theta'$. Angular velocity $\dot{\theta}$ is then well estimated by the kPOMDPs. The rightmost figure shows the averaged result of the total rewards obtained by learned policies as training samples increased. More description is in the text.

posterior embedding in (10),

$$\hat{\mu}_{\bar{X}|y} = \hat{C}_{\bar{X}\bar{Y}} \left( \hat{C}^2_{\bar{Y}\bar{Y}} + \delta n I_n \right)^{-1} \hat{C}_{\bar{Y}\bar{Y}} k_{\mathcal{Y}}(y, \cdot), \qquad (22)$$

where $\delta$ is a small regularization parameter to avoid the instability of $\hat{C}^{-1}_{\bar{Y}\bar{Y}}$, since the empirical estimate $\hat{C}_{\bar{Y}\bar{Y}}$ may include negative weights. The estimate (22) is consistent under smoothness assumptions, and converges to $\mu_{\bar{X}|y}$ in the infinite sample limit. Since we always normalize weight vectors to probability vectors as in subsection 4.2, however, we use the simpler and more computationally efficient estimate

$$\hat{\mu}_{\bar{X}|y} = \hat{C}_{\bar{X}\bar{Y}} (\hat{C}_{\bar{Y}\bar{Y}} + \epsilon n I_n)^{-1} k_{\mathcal{Y}}(y, \cdot), \qquad (23)$$

where $\epsilon$ is a small regularization parameter. Though the consistency of (23) with the combination of the normalization is not theoretically proven, experiments (Section 5) show good empirical results. In what follows we give the KBR algorithm when the estimate (23) is used with normalized weights.

Denote feature vectors $\varphi(X) = k_{\mathcal{X}}(X, \cdot)$ and $\phi(Y) = k_{\mathcal{Y}}(Y, \cdot)$. Let $U_1, \ldots, U_l$ be $l$ samples drawn from the prior $\Pi$ and $(X_1, Y_1), \ldots, (X_n, Y_n)$ be $n$ samples drawn from $P$. Consider an empirical prior embedding $\hat{\mu}_\Pi = \bar{\Upsilon}\boldsymbol{\gamma}$ where $\bar{\Upsilon} = (\varphi(U_1), \ldots, \varphi(U_l))$ are feature mappings of the prior samples and $\boldsymbol{\gamma} \in \mathbb{R}^l$ is a weight vector. Define the $n \times l$ matrix $G_{XU} = \Upsilon^\top \bar{\Upsilon}$, and

$$\Upsilon \circ \Phi := (\varphi(X_1) \otimes \phi(Y_1), \ldots, \varphi(X_n) \otimes \phi(Y_n)),$$
$$\Phi \circ \Phi := (\phi(Y_1) \otimes \phi(Y_1), \ldots, \phi(Y_n) \otimes \phi(Y_n)). \qquad (24)$$

Empirical estimates of the covariance operators $C_{\bar{X}\bar{Y}}$, $C_{\bar{Y}\bar{Y}}$ are then given by $\hat{C}_{\bar{X}\bar{Y}} = (\Upsilon \circ \Phi)\boldsymbol{\beta}$, $\hat{C}_{\bar{Y}\bar{Y}} = (\Phi \circ \Phi)\boldsymbol{\beta}$ with the weight vector $\boldsymbol{\beta} = (G_X + \varepsilon n I_n)^{-1} G_{XU}\boldsymbol{\gamma}$ [Fukumizu et al., 2011]. We approximate $\boldsymbol{\beta}$ by a non-negative vector $\hat{\boldsymbol{\beta}}$, as in subsection 4.2, leading to the following proposition:

**Proposition 1.** *The empirical estimate eq.(23) has the following Gram matrix expression using a non-negative vector $\hat{\boldsymbol{\beta}}$ for all $y \in \mathcal{Y}$:*

$$\hat{\mu}_{\bar{X}|y} = \Upsilon R_{X|Y}(\hat{\boldsymbol{\beta}}) \mathbf{k}_Y(y),$$
$$R_{X|Y}(\hat{\boldsymbol{\beta}}) := \left( D(\hat{\boldsymbol{\beta}}) G_Y + \delta n I_n \right)^{-1} D(\hat{\boldsymbol{\beta}}), \qquad (25)$$

*where $\mathbf{k}_Y(y) = \Phi^\top \phi(y)$ and $D(\hat{\boldsymbol{\beta}}) = \mathrm{diag}(\hat{\boldsymbol{\beta}})$.*

*Proof.* The proof follows the same reasoning as Proposition 3.4 of [Fukumizu et al., 2011]. □

Let $\boldsymbol{\alpha}(y) = R_{X|Y}(\boldsymbol{\beta})\Phi^\top \phi(y) \in \mathbb{R}^n$. The posterior embedding $\mu_{\bar{X}|y}$ can be estimated by a linear combination of feature vectors on samples $\Upsilon = (\varphi(X_1), \ldots, \varphi(X_n))$ with weights $\boldsymbol{\alpha}(y) \in \mathbb{R}^n$ that depend on $y \in \mathcal{Y}$.

**Acknowledgment** This work has been supported in part by JSPS KAKENHI (B) 22300098. Authors thank anonymous reviewers for helpful comments.

# References

[Even-dar, 2005] Eyal Even-dar. Reinforcement learning in POMDPs without resets. *IJCAI*, 690–695, 2005.

[Fine and Scheinberg, 2001] S. Fine and K. Scheinberg. Efficient SVM training using low-rank kernel representations. *JMLR*, 2:243–264, 2001.

[Fukumizu et al., 2008] K. Fukumizu, A. Gretton, X. Sun, and B. Schölkopf. Kernel measures of conditional dependence. In *NIPS2008*.

[Fukumizu et al., 2011] K. Fukumizu, L. Song, and A. Gretton. Kernel Bayes' Rule. In *NIPS2011*.

[Fukumizu et al., 2011] K. Fukumizu, L. Song, and A. Gretton. Kernel bayes rule: Bayesian inference with positive definite kernels. *arXiv:1009.5736*.

[Gretton et al., 2007] A. Gretton, K. Borgwardt, M. Rasch, B. Schölkopf, and A. Smola. A kernel method for the two-sample-problem. In *NIPS2007*.

[Gretton et al., 2008] A. Gretton, K. Fukumizu, C. Teo, L. Song, B. Schölkopf, and A. Smola. A kernel statistical test of independence. In *NIPS2008*.

[Gretton et al., 2012] A. Gretton, K. Borgwardt, M. Rasch, B. Schölkopf and A. Smola. A Kernel Two-Sample Test. *JMLR*, 13, 671–721, 2012.

[Grunewalder et al., 2012] S. Grunewalder, G. Lever, L. Baldassarre, M. Pontil and A. Gretton. Modelling transition dynamics in MDPs with RKHS embeddings. In *ICML2012*.

[Hauskrecht, 2000] M. Hauskrecht. Value-Function Approximations for Partially Observable Markov Decision Processes. In *JAIR*, vol 13, pages 33–94, 2000.

[Littman, 1995] M. Littman, A. Cassandra, and L. Kaelbling. Learning policies for partially observable environments: Scaling up. In *ICML1995*.

[Pineau et al., 2003] J. Pineau, G. Gordon, and S. Thrun. Point-based value iteration: an anytime algorithm for POMDPs. In *ICJAI*, pages 1025–1032, 2003.

[Porta. et al., 2006] J. M. Porta, N. Vlassis, and P. Poupart. Point-based value iteration for continuous POMDPs. *JMLR*, 7:2329–2367, 2006.

[Poupart et al., 2006] Pascal Poupart, Nikos A. Vlassis, Jesse Hoey, and Kevin Regan. An analytic solution to discrete Bayesian reinforcement learning. *ICML2006*.

[Ross et al., 2008] S. Ross, J. Pineau, S. Paquet, and B. Chaib-draa. Online planning algorithms for POMDPs. *JAIR*, 32(1):663–704, 2008.

[Silver and Veness, 2010] David Silver and Joel Veness. Monte-Carlo Planning in Large POMDPs. *NIPS2010*.

[Smith and Simmons, 2004] T. Smith and R. Simmons. Heuristic search value iteration for POMDPs. In *UAI2004*.

[Smola et al., 2007] A. Smola, A. Gretton, L. Song, and B. Schölkopf. A Hilbert space embedding for distributions. In *ALT2007*.

[Sondik, 1971] E. J. Sondik. *The Optimal Control of Partially Observable Markov Processes*. PhD thesis, Stanford University, 1971.

[Song et al., 2009] L. Song, J. Huang, A. Smola, and K. Fukumizu. Hilbert space embeddings of conditional distributions with applications to dynamical systems. In *ICML2009*.

[Song et al., 2010] L. Song, B. Boots, S. Siddiqi, G. Gordon, and A. Smola. Hilbert space embeddings of hidden Markov models. In *ICML2010*.

[Song et al., 2010] L. Song, A. Gretton, and C. Guestrin. Nonparametric tree graphical models via kernel embeddings. In *AISTATS*, pages 765–772, 2010.

[Song et al., 2011] L. Song, A. Gretton, D. Bickson, Y. Low, and C. Guestrin. Kernel Belief Propagation. In *AISTATS*, 2011.

[Spaan and Vlassis, 2005] M. T. J. Spaan and N. Vlassis. Perseus: Randomized point-based value iteration for POMDPs. *JAIR*, 24:195–220, 2005.

[Sriperumbudur et al., 2010] B. Sriperumbudur, A. Gretton, K. Fukumizu, G. Lanckriet, and B. Schölkopf. Hilbert space embeddings and metrics on probability measures. *JMLR*, 11:1517–1561, 2010.

[Thrun, 2000] S. Thrun. Monte Carlo POMDPs. In *NIPS2000*.